%% file: main.tex
\documentclass[10pt,twocolumn,letterpaper]{article}

\usepackage{multirow}
\usepackage{graphicx}
\usepackage[pagenumbers]{cvpr} 

\input{preamble}

\definecolor{tablegray}{gray}{0.9}
\definecolor{cvprblue}{rgb}{0.21,0.49,0.74}
\usepackage[pagebackref,breaklinks,colorlinks,allcolors=cvprblue]{hyperref}

\title{GLUS: Global-Local Reasoning Unified into \\ A Single Large Language Model for Video Segmentation}

\author{Lang Lin$^{*}$\quad\quad Xueyang Yu$^{*}$\quad\quad Ziqi Pang$^{*}$ \quad\quad Yu-Xiong Wang \\
University of Illinois Urbana-Champaign}

\begin{document}
\twocolumn[{%
\renewcommand\twocolumn[1][]{#1}%
\maketitle
\input{sec/0_0_teaser}
}]
\begin{NoHyper}
\def\thefootnote{*}\footnotetext{Equal Contribution.}
\end{NoHyper}
\def\thefootnote{\arabic{footnote}}
\input{sec/0_abstract}    
\input{sec/1_intro}
\input{sec/2_related}
\input{sec/3_method}
\input{sec/4_exp}
\input{sec/5_conclusion}
{
    \small
    \bibliographystyle{ieeenat_fullname}
    \bibliography{main}
}
\clearpage
\input{sec/X_suppl}
\end{document}

%% file: preamble.tex
\newcommand\mypar[1]{\par\vspace{1.0mm}\noindent\textbf{#1}\;\;}

\newcommand{\ourwork}{GLUS\xspace}

\expandafter\def\expandafter\normalsize\expandafter{%
    \normalsize%
    \setlength\abovedisplayskip{3pt}%
    \setlength\belowdisplayskip{3pt}%
    \setlength\abovedisplayshortskip{-3pt}%
    \setlength\belowdisplayshortskip{3pt}%
}

\newcommand\seg{\left<\mathtt{SEG}\right>}
\newcommand\score{\left<\mathtt{SCORE}\right>}
\usepackage{pifont}
\newcommand{\cmark}{\ding{51}}%
\newcommand{\xmark}{\ding{55}}%

%% file: sec/0_0_teaser.tex
\begin{center}
    \centering
    \captionsetup{type=figure}
    \includegraphics[width=0.99\textwidth]{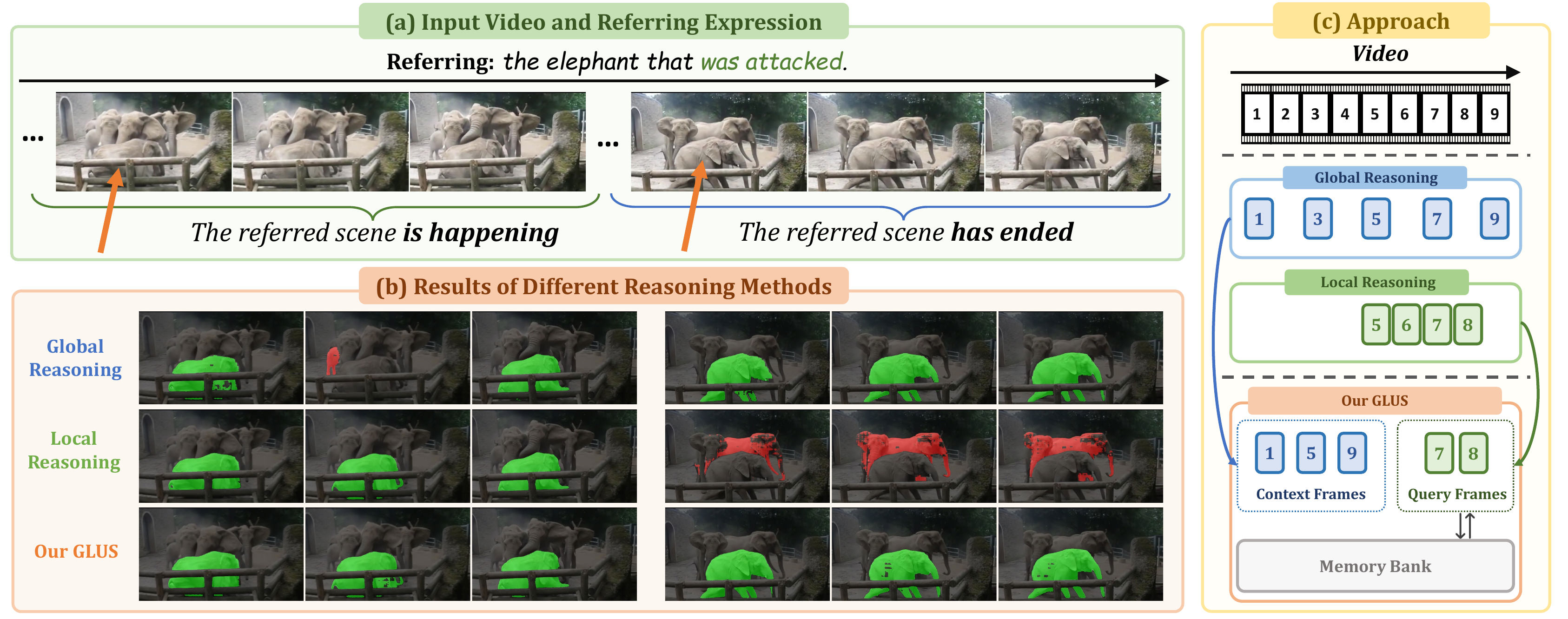}
    \vspace{-2mm}
    \captionof{figure}{Overview of \textbf{\ourwork}. \textbf{(a)} We present a representative RefVOS example where the referring expression only covers part of the frames (the \textcolor{orange}{orange} arrows point to the ground truth). Therefore, RefVOS requires both \emph{global reasoning} for finding the target objects in frames without the referred motions and \emph{local reasoning} for predicting temporally consistent masks. \textbf{(b)} We show \textcolor{green}{green} and \textcolor{red}{red} masks for correct and wrong segmentation. Existing multimodal large language models (MLLMs) focus on either global or local reasoning. However, global-only shows fluctuation in local ``VOS,'' while local-only ``Ref'' to wrong objects without overall video contexts when the referring is not happening. Instead, our unified global-local reasoning shows consistent and correct predictions. \textbf{(c)} \ourwork provides a simple yet effective baseline that unifies global and local reasoning through both global and local frame sampling and end-to-end memory modules. 
    } 
    \label{fig:teaser}
\end{center}

%% file: sec/0_abstract.tex
\begin{abstract}

This paper proposes a novel framework utilizing multi-modal large language models (MLLMs) for referring video object segmentation (RefVOS). Previous MLLM-based methods commonly struggle with the dilemma between ``Ref'' and ``VOS'': they either specialize in understanding a few key frames (global reasoning) or tracking objects on continuous frames (local reasoning), and rely on external VOS or frame selectors to mitigate the other end of the challenge. However, our framework \textbf{GLUS} shows that \textbf{G}lobal and \textbf{L}ocal consistency can be \textbf{U}nified into a single video \textbf{S}egmentation MLLM: a set of sparse ``context frames'' provides global information, while a stream of continuous ``query frames'' conducts local object tracking. This is further supported by jointly training the MLLM with a pre-trained VOS memory bank to simultaneously digest short-range and long-range temporal information. To improve the information efficiency within the limited context window of MLLMs, we introduce object contrastive learning to distinguish hard false-positive objects and a self-refined framework to identify crucial frames and perform propagation. By collectively integrating these insights, our \textbf{GLUS} delivers a simple yet effective baseline, achieving new state-of-the-art for MLLMs on the MeViS and Ref-Youtube-VOS benchmark. Our project page is at \href{https://glus-video.github.io/}{https://glus-video.github.io/}.

\end{abstract}

%% file: sec/1_intro.tex
\vspace{-4mm}
\section{Introduction}
\label{sec:intro}

Using language expressions to specify target objects requires joint reasoning of visual contents and language intentions. Such capabilities in videos are recently studied via the task of referring video object segmentation (RefVOS), which requires the models to localize and consistently track target object(s) according to language descriptions, such as ``\emph{the elephant that was attacked}'' in Fig.~\ref{fig:teaser}\textcolor{cvprblue}{(a)}. 
The unique challenge of RefVOS is that the described action may only be observable in a subset of frames.
Therefore, the models must understand the global characteristics of each object to reliably propagate such reasoning across the whole video.

Motivated by the reasoning capabilities of multi-modal large language models (MLLMs) in referring image segmentation (RIS), \emph{e.g.}, LISA~\cite{lai2024lisa}, RefVOS studies~\cite{yan2024visa, zheng2024villa, bai2024one} adapt these MLLMs to videos, hoping to reason the variety of language expressions effectively. These approaches commonly consider MLLM as a \emph{multi-image} framework and reason a limited number of $N$ frames. Consequently, they struggle to handle the entire video, especially with their context window $N$ significantly fewer than the video frames to process. These MLLMs then face the shared ``Ref'' and ``VOS'' dilemma in RefVOS: using the $N$ frames to either understand the \emph{global} context or address \emph{local} temporal continuity. Focusing solely on either global or local aspects results in inadequate performance for complex RefVOS scenarios (as in Fig.~\ref{fig:teaser}\textcolor{cvprblue}{(b)}). To satisfy both aspects, they resort to \emph{external} modules like off-the-shelf video object segmentation (VOS) models, which unavoidably increases the system's complexity. Instead, we aim to demonstrate a framework \textbf{\ourwork} where \emph{a single MLLM alone has the ability of \textbf{G}lobal-\textbf{L}ocal \textbf{U}nified reasoning for video \textbf{S}egmentation}.

Our \ourwork begins from a simple yet principled adaptation to MLLM by \emph{providing} and \emph{only providing} the necessary information for \emph{global} and \emph{local} reasoning. For global reasoning, the MLLM must have a set of frames covering the whole video to maximize the probability of observing the referred property. For local reasoning, we realize that MLLMs are natively compatible with the VOS formulation, because the current predictions are always based on the precedent frames. Therefore, our \ourwork divides the $N$ frames for an MLLM into two groups (as in Fig.~\ref{fig:teaser}\textcolor{cvprblue}{(c)}): \textbf{(1) Context frames}: several frames uniformly sampled from the video to cover the global reasoning context; \textbf{(2) Query frames}: the frames to produce segmentation results, which are continuously sampled to support temporal continuity naturally. Such a procedure also intuitively mimics the behaviors of a human to address video segmentation: first, check a few sparsely sampled frames (\emph{i.e.}, context frames) to understand the video; then, drawing the masks of an object on frames iteratively (\emph{i.e.}, query frames). 

Compared with global-only and local-only strategies, \ourwork divides total $N$ frames into two types and inevitably reduces both global and local video information obtained from the video. To tackle the issue, we seek to break the limits of an MLLM's context window size $N$ by introducing a memory bank that can store information from previous predictions, which enhances both local and global reasoning. Since our query frames are continuous, we discover that \ourwork can be trained and inferred jointly with the memory module from a pre-trained foundational VOS model, \emph{e.g.}, SAM2~\cite{ravi2024sam2}. By creatively unifying such an ``\emph{online}'' VOS module and optimizing it end-to-end with an MLLM, our \ourwork offers a more straightforward system without calling external VOS models.  

Enabled by our framework capturing global and local contexts, which provides more comprehensive temporal contexts than conventional approaches, we can better explore \emph{distinguishing the target information in a fine-grained way}. From the frame level, the prediction quality of \ourwork can reflect the relevance between a frame and referring expression, and create useful pseudo-labels for video MLLMs to understand the notion of ``key frames.'' Contrasting the previous works\cite{yan2024visa} using \emph{off-the-shelf} key frame selectors, our fine-tuning enables better contextual information for our global-local reasoning. The selector trained from \ourwork is further used for internal propagation which leads to a self-refinement framework. From the object level, \ourwork enhances the fine-grained matching between objects and language instructions via contrastive losses, maximizing the distance of tokens referring to different objects. 

In conclusion, our contributions in \ourwork are:
\begin{enumerate}
    \item We demonstrate that unifying global and local reasoning into a \emph{single} MLLM for RefVOS through the design of \emph{context} and \emph{query} frames constitutes a simple yet effective baseline method for MLLM-based RefVOS models.
    \item We illustrate \emph{end-to-end training of a MLLM with pre-trained VOS memory modules} and decouple the needs for calling external VOS models.
    \item We introduce plug-and-play self-refinement with key frame selectors and object contrastive loss distinguishing hard false-positive objects.
\end{enumerate}

Our simple yet effective \ourwork serve as a strong MLLM-based RefVOS baseline, demonstrating advantages on the major RefVOS benchmarks MeViS~\cite{ding2023mevis} and Ref-Youtube-VOS~\cite{seo2020urvos} over previous MLLMs.

%% file: sec/2_related.tex
\input{tables/formulation_comparison}
\vspace{-2mm}
\section{Related Work}
\label{sec:related}

\vspace{-2mm}
\mypar{Referring Video Object Segmentation.} 
Referring video object segmentation (RefVOS) aims to segment the target object within a video based on a given language expression. A recent benchmark, MeViS~\cite{ding2023mevis}, introduces complex multi-object scenarios with extensive motion dynamics, posing greater challenges to the community. Common practices~\cite{wu2023onlinerefer, SOC, tang2023temporal, yuan2024losh, wu2022referformer, mttr} utilize language queries to attend to the referred object. Some recent works ~\cite{ding2023mevis, DsHmp} propose motion aggregation to capture motion information. Besides, recent works~\cite{zheng2024villa, yan2024visa, bai2024one, zhu2023tracking, munasinghe2024videoglamm} use multimodal large language models (MLLMs) to reason over complex expressions. In this work, we propose a novel MLLM-based framework that unifies both global and local reasoning and decouples the need of motion aggregation modules.

\mypar{Large Language Models in Segmentation and Grounding.}
Inspired by the impressive reasoning capabilities of large language models (LLMs), researchers are seeking to equip LLMs with fine-grained understanding in visual tasks ~\cite{lai2024lisa, zhang2024groundhog, ren2024pixellm, munasinghe2023pg, pi2023detgpt}.  LISA~\cite{lai2024lisa} pioneered such approach by connecting a multi-modal LLM (MLLM) with the Segment Anything Model~\cite{kirillov2023segany} via a special token to produce accurate segmentation masks. Some recent works extend LISA~\cite{lai2024lisa} to the video domain. TrackGPT~\cite{zhu2023tracking} makes a straightforward adaptation by iteratively updating the special token as the video progresses. VISA~\cite{yan2024visa} further provides global context for producing a special token, while ViLLa~\cite{zheng2024villa} designs a context-aggregation module to derive more meaningful visual features. VideoLISA~\cite{bai2024one} uses a single token for tracking throughout the video. In this work, we introduce an effective MLLM baseline to endow the model with better motion understanding abilities without additional VOS modules needed.

\mypar{Video Object Segmentation.} Different from RefVOS, VOS is an \emph{online} task: the target object is marked with a ground truth mask in the first frame, and the VOS models function in a streaming way to track this object. The memory bank, which stores historical information, is the core module that enables the temporal reasoning abilities of VOS. Existing methods commonly leverage pixel-level~\cite{chen2018blazingly, hu2018videomatch, yang2018efficient, yang2020collaborative, yang2021collaborative, voigtlaender2019feelvos} or object-level~\cite{athar2022hodor, athar2023tarvis, cheng2023putting, li2024univs} matching via attention mechanisms~\cite{vaswani2017attention} when using memory banks. In addition to interacting with memory banks, VOS studies~\cite{cheng2022xmem, zhou2024rmem, ravi2024sam2, yang2024samurai}  also discover the importance of organizing the memory banks properly. In the context of RefVOS, our framework \ourwork supports the benefits of training a pre-trained memory bank end-to-end with a RefVOS MLLM, including both its memory reading attention and organization. This enables the MLLM to reason with information beyond its own context window and acquire the ability of consistent object tracking.

%% file: tables/formulation_comparison.tex
\begin{table*}[t]
\centering
\resizebox{0.9\textwidth}{!}{
\begin{tabular}{l@{\hspace{8mm}}|c@{\hspace{6mm}}c@{\hspace{6mm}}|c@{\hspace{6mm}}c@{\hspace{6mm}}|c|c}
\toprule
 & \multicolumn{2}{c|}{Training} & \multicolumn{2}{c|}{Inference} & \multirow{2}{*}{$\seg$} & \multirow{2}{*}{External VOS} \\ 
 & Sampled Frames & Information & Sampled Frames & Information & & \\
 \midrule
VideoLISA~\cite{bai2024one} & Uniform & Global & Uniform  & Global & 1 & \cmark \\
VISA~\cite{yan2024visa} & Random & Random & Uniform + Continuous & Global + Local & 1 & \cmark \\
ViLLa~\cite{zheng2024villa} & Continuous & Local & Continuous & Local & $N$ & \xmark \\
\midrule
\ourwork (Ours) & Uniform + Continuous & Global + Local & Uniform + Continuous & Global + Local & $N$ & \xmark \\
\bottomrule
\end{tabular}
}
\vspace{-3mm}
\caption{Comparison of \emph{frame} and \emph{information} utilized in existing RefVOS MLLMs. Despite the distinct challenges of ``Ref'' (global reasoning) and ``VOS'' (local reasoning) in RefVOS, previous methods fail to unify them and offer inconsistent strategies for training and inference. Compared with previous methods, our \ourwork proposes the ``context + query frames'' strategy (Sec.~\ref{sec:method}) that effectively unifies global and local reasoning for both training and inference stages. 
}
\vspace{-5mm}
\label{tab:formulation}
\end{table*}

%% file: sec/3_method.tex
\section{Preliminaries}
\label{sec:prelim}

\subsection{Referring Video Object Segmentation} 

Given an input video consisting of $T$ frames $I_{1:T} \in \mathbb{R}^{T\times H\times W\times 3}$ and a referring language expression $R$, RefVOS aims to build a model $\phi_\theta$ predicting binary segmentation masks $\mathcal{M}_{1:T}$ of the referred object:
\begin{equation}\label{eqn:refvos}
    M_{1:T} = \phi_\theta(I_{1:T}, R)
\end{equation}

\begin{figure*}[t]
    \centering
    \includegraphics[width=0.99\textwidth]{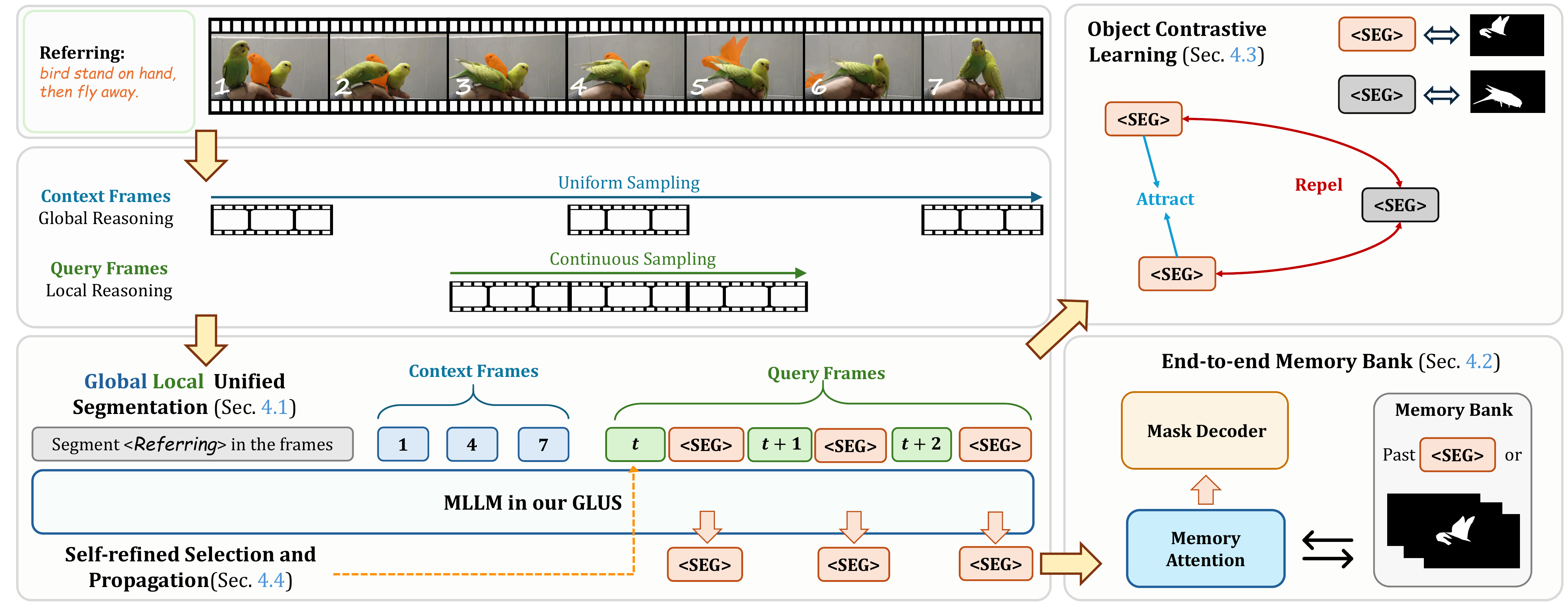}
    \vspace{-3mm}
    \caption{Overview of our \ourwork. The yellow arrows mark the flow. \textbf{(1)} Sec.~\ref{sec:global-local-segmentor}. Beginning from a video and referring, we design \emph{context frames} and \emph{query frames} to unify the distinct global and local reasoning into a single MLLM. The $\seg$ tokens represent the target object. \textbf{(2)} Sec.~\ref{sec:memory_bank}. \ourwork end-to-end integrates pre-trained VOS memory modules to enhance temporal reasoning and decouple the reliance of existing models on VOS post-processing. The memory-enhanced decoder decodes $\seg$ tokens into masks. \textbf{(3)} Sec.~\ref{sec:contrastive-loss}. The $\seg$ tokens are further supervised to better distinguish the different objects. \textbf{(4)} Sec.~\ref{sec:selector}. The accuracy of $\seg$ can be used for finetuning a plug-and-play key frame selector to further enhance \ourwork's inference-time ability. 
    }
    \label{fig:pipeline}
    \vspace{-3mm}
\end{figure*}

RefVOS differs from both VOS and referring image segmentation (RIS) tasks. Compared with VOS, RefVOS is an offline task. It exhibits the critical challenge of localizing the target object and action from the whole video, where only a short key video clip might correspond to the language expression (as Fig.~\ref{fig:pipeline}). Compared with RIS, RefVOS requires global video reasoning and temporal coherence with consistent objects across numerous frames.

\subsection{LISA: MLLM for Segmentation}  

As matching target objects with language expressions demands reasoning capabilities, recent MLLM-based RefVOS methods mainly follow the successful image-based segmentation models represented by LISA~\cite{lai2024lisa}. LISA designs a dedicated $\seg$ token to represent the target object and call a segmentation decoder $\mathtt{Dec}$~\cite{kirillov2023segany} to predict the object masks $M$:
\begin{equation}\label{eqn:lisa}
    \seg = \mathtt{MLLM}(I, R), M = \mathtt{Dec}(I, \seg),
\end{equation}
where we slightly simplify the MLLM to output the segmentation token only. Besides LISA, more studies~\cite{zhang2024groundhog, rasheed2024glamm} also enable MLLMs for RIS and object grounding tasks, but we mainly discuss LISA since the majority of MLLMs in RefVOS~\cite{bai2024one, yan2024visa, zheng2024villa} follow its design.

\subsection{Adapting Image MLLMs for RefVOS} 
\label{sec:prelim_discuss}

To adapt an image-based framework like LISA to videos, the natural intuition is to extend it into a \emph{multi-image} framework. Concretely, the MLLM has to take multiple frames $I_{1:T}$ as input, and the segmentation token(s) is used to decode masks for each frame, as adapted from Eqn.~\ref{eqn:lisa}: 
\begin{equation}\label{eqn:lisa_refvos}
\begin{split}
    & \seg_{1:N} = \mathtt{MLLM}(I_{1:N}, R), \\ & M_{1:N} = \mathtt{Dec}(I_{1:N}, \seg_{1:N}),
\end{split}
\end{equation}
where $N$ is the maximum number of frames an MLLM can take (\emph{e.g.}, 16) for training. As $N$ is commonly smaller than the total number of video frames $T$ that require segmentation, we observe different strategies for bridging this gap from the aspects listed in Table~\ref{tab:formulation}. These methods mostly sample $N$ key frames and utilize an external VOS model~\cite{cheng2022xmem} to propagate the masks. As discussed in Sec.~\ref{sec:intro}, this not only makes the system more complicated but also restricts the propagation from using the language instructions as guidance. Moreover, the performance of the MLLM is also bounded by the key frame selector, which is yet another multi-modal understanding task. 

\section{Methods}
\label{sec:method}

Our \ourwork addresses these challenges via unifying global and local reasoning into a single MLLM (Sec.~\ref{sec:global-local-segmentor}). Then we demonstrate an end-to-end framework involving both MLLM and a pre-trained VOS memory bank, which significantly enhances the temporal continuity of our framework and decouples the reliance on calling an off-the-shelf VOS model (Sec.~\ref{sec:memory_bank}). To further improve the fine-grained matching between language expressions and objects in global reasoning, we introduce an object-level contrastive loss (Sec.~\ref{sec:contrastive-loss}). Finally, we discuss the potential of our \ourwork in enhancing the key frame selectors, which are commonly used as off-the-shelf oracles in prior arts~\cite{yan2024visa, bai2024one}, via producing the pseudo-labels for finetuning (Sec.~\ref{sec:selector}). An overview of \ourwork is in Fig.~\ref{fig:pipeline}. 

\subsection{Global-Local Unified MLLM}
\label{sec:global-local-segmentor}

\mypar{Dilemma of Global and Local Reasoning for Conventional MLLMs.} The task of ``RefVOS'' involves two distinct abilities of global and local reasoning as described in Sec.~\ref{sec:prelim_discuss}. However, unifying them is contradicting for a single MLLM. \textbf{(1)} \emph{Global reasoning}. As a most straightforward adaptation from LISA~\cite{lai2024lisa}, the input to the MLLM is $N$ uniformly sampled frames of the video to capture global contexts. The output is then $N$ or 1 corresponding $\seg$ tokens for these frames. \textbf{(2)} \emph{Local reasoning}. Although previous methods~\cite{yan2024visa, bai2024one} utilize external VOS for local mask propagation, we realize that the autoregressive formulation of MLLMs is already sufficient to address temporal consistency, where the token $\seg_{t}$ is generated by using previous predictions $\seg_{1:t-1}$ as contexts. However, the limitation of local reasoning is the adjacency of the $N$ frames. 

Previous approaches cannot unify both aspects because their training and inference styles are fundamentally different: models emphasizing global reasoning have to divide the video into separate folds to cover long-range contexts. In contrast, local reasoning models mainly rely on a sliding window covering a short range of neighboring frames. As in Table~\ref{tab:formulation}, previous methods deal with either global or local reasoning when training the models.

\mypar{Context and Query Frames.} 
Different from previous methods, our \ourwork bypasses the above dilemma by explicitly unifying both global and local contexts for a single MLLM as in Fig.~\ref{fig:pipeline}. Regarding global reasoning, we propose a set of sparse ``\emph{context frames}'' $I^C_{1:N_C}$ covering the overall context of the videos and supporting the matching between language instructions and objects. Then \ourwork introduces a set of continuous ``\emph{query frames}'' $I^Q_{1:N_Q}$ for local reasoning and decoding the segmentation masks. As in Fig.~\ref{fig:pipeline}, our \ourwork places the query frames after the context frames because MLLMs function in a causal direction and such an order enables the decoding of segmentation results to receive video contexts. Intuitively, we imitate how a human would address the RefVOS task: first checking the video contents roughly to localize an object and then marking the object locations iteratively on every frame.

Formally, we delve deep into the LLM part of our MLLM, and it generates the $t$-th segmentation token as: 

\begin{equation}
\label{eqn:unified_pred}
\begin{split}
    \seg_t = \mathtt{LLM}([& R, I^C_{1:N_c}, \\
                           & I^Q_{1}, \seg_1, I^Q_{2}, \seg_2, ..., I^Q_{t}]),
\end{split}
\end{equation}
where we autoregressively apply the referring, context frames, precedent query frames and segmentation tokens as the context for prediction. With such a design, our method \ourwork unifies both global and local understanding in a simple and training-inference consistent way, which overcomes the limitations of previous MLLMs (Table~\ref{tab:formulation}).

\mypar{Training.} We adopt the most straightforward way of preparing context frames and query frames. Regarding context frames $I^C_{1:N_C}$, we first split the entire video into $N_C$ equally spaced clips and then randomly sample one frame from each video clip to obtain $N_c$ context frames, which is similar to~\cite{wang2016temporal}. For the query frames, we randomly sample short clips of $N_Q$ frames to imitate the procedure of iteratively decoding masks for every frame of a video.

Our loss function is similar to the original LISA~\cite{lai2024lisa} involving both text and mask supervision. The objective of text supervision is enforcing the underlying LLM to generate the special $\seg$ token and our \ourwork adopts the standard cross entropy loss for this. Regarding mask supervision, we follow SAM2~\cite{ravi2024sam2} in combining the per-pixel binary cross-entropy (BCE) loss and DICE loss. More details are in the supplementary materials Sec ~\ref{sec:glus_training} .

\mypar{Inference.} Our \ourwork exhibits a fully aligned inference procedure with training, which is a significant advantage. For context frames, we first evenly divide the video into $N_C$ clips then select their center frame as context frames. Such a set of uniformly sampled frames cover the overall video contexts for global reasoning. Please note that this set of context frame remains identical during the whole inference procedure on this video. To obtain the masks for all of the $T$ frames in the video, we use sliding windows with size $N_q$ and stride $1$ to generate a batch of query frames groups. With the frames being adjacent to each other, our strategy maximizes the temporal continuity for local reasoning.

\subsection{End-to-end Memory Banks for MLLMs} 
\label{sec:memory_bank}

\mypar{Motivation: Rethinking VOS Models for RefVOS.} To address the local continuity problem, existing MLLMs~\cite{bai2024one, yan2024visa} commonly treat VOS as an external module. One of the advantages of VOS methods compared with MLLMs is their using a memory bank to store long-term historical information, which is usually larger than the context window $N$ of an MLLM. In addition, the VOS models also involve specialized memory reading and updating operations~\cite{cheng2022xmem, ravi2024sam2} empowered by pre-trained transformers. Therefore, our \ourwork aims to end-to-end unify memory banks into MLLMs to enhance the ability of MLLMs for temporal reasoning and simplify the RefVOS framework. This significantly enhances \ourwork in maintaining long-range temporal information for global and local reasoning.

\mypar{Design.} The key principle of our unified memory bank is the joint optimization of the memory bank modules and MLLM end-to-end as in Fig.~\ref{fig:pipeline}. Concretely, the decoding of the $t$-th query frame further involves the memory bank:
\begin{equation}
\label{eqn:mem_bank}
M_t = \mathtt{Dec}(I^Q_t,\seg_t, \mathtt{MemBank}).
\end{equation}
Therefore, the gradient can be back-propagated from both the features stored in the memory banks and the pre-trained VOS attention used to read the memory bank features. In this way, our MLLM can cooperate with pre-trained memory banks from foundational VOS models and enjoy their enhancement in attending to historical information.

Our joint optimization enables aligned training and inference distributions, different from calling an external VOS model during inference time. In our case, the memory bank iterates through all the video frames with our query frames $I_{t}^Q$ predicting the masks frame by frame. Accordingly, our training simulates such a streaming behavior with the $N_Q$ query frames. As later verified in Sec.~\ref{sec:exp}, our memory bank is a convenient \textit{plug-and-play} component enhancing the ``VOS'' ability of RefVOS MLLMs. Its effectiveness further supports our design of dividing the frames into global context and local reasoning ones, which can seamlessly enjoy the progress in VOS models.

\subsection{Object Contrastive Loss}
\label{sec:contrastive-loss}

\mypar{Motivation.} In addition to unifying global and local reasoning from an architectural perspective, we advance our investigation to enhance the correct matching between language instructions and target objects, which is the critical challenge of referring segmentation tasks. 

As shown in the example of Fig.~\ref{fig:teaser} and Fig.~\ref{fig:pipeline}, a video might contain multiple objects with similar appearances to the ground truth. In this case, MLLM may easily confuse these objects and generate similar $\seg$ tokens. Therefore, we aim to enhance the fine-grained perception of MLLMs by distinguishing such confusing object pairs.

\mypar{Design.} We introduce the object contrastive loss that maximizes the distance between the $\seg$ tokens referring to different objects. When constructing positive pairs, \ourwork notices the uniqueness of MeViS in that multiple different language expressions might refer to the same object. Therefore, the $\seg$ tokens generated from different referring expressions of the same object are naturally positive samples. For negative sample pairs, \ourwork aims to construct a sufficient number of negative examples motivated by~\cite{chen2020simple}. This is achieved by maintaining a segmentation token bank \cite{he2020momentum, he2024decoupling, wu2022defense, el2023learning} of different objects. 

Formally, the object contrastive loss follows the SimCLR~\cite{chen2020simple}, where the feature $\mathbf{v}$ of a $\seg$ token is supervised by both positive sample $\mathbf{k}^{+}$ and negative samples $\mathbf{k}^{-}$:

\begin{small}
\begin{equation}
    \begin{aligned}
        -\sum_{\mathbf{v}, \mathbf{k}^{+}} \log \frac{\exp \left(\operatorname{sim}(\mathbf{v}, \mathbf{k}^{+}) / \tau\right)}{\exp \left(\operatorname{sim}(\mathbf{v}, \mathbf{k}^{+}) / \tau\right) + \sum_{\mathbf{k}^{-}} \exp (\operatorname{sim}\left(\mathbf{v}, \mathbf{k}^{-}\right)/\tau)}. 
    \end{aligned}
\end{equation}
\end{small}

However, the natural challenge of contrastive loss is that not all objects have multiple distinct referring expressions and, accordingly, positive sample pairs. Thus, the contrastive loss is computed only when positive samples are presented. According to our statistics, the likelihood of sampling positive pairs within a data batch is approximately 91.5\% for MeViS. As Ref-Youtube-VOS relatively lacks multiple complex referring expressions for the same objects, our contrastive loss is only applied to MeViS samples, but surprisingly benefits Ref-Youtube-VOS as well. 

\subsection{Self-refined Selection and Propagation} 
\label{sec:selector}

\mypar{Guiding Key Frame Selection with \ourwork.} As the language instructions might refer to a motion occurring on parts of the videos, RefVOS methods commonly select the key frames to assist the localization of target objects. Although \ourwork has already shown effective global-local unified reasoning, we are limited by the context window of MLLMs, and the sparsely sampled context frames could miss crucial information. Existing methods~\cite{yan2024visa} adopt off-the-shelf video QA models for key frame selection, but such models are not aligned with the ``key frame selection'' objective for RefVOS models. In this sense, we suggest that the prediction quality from our \ourwork is a natural indicator of the relevance between language instructions and video frames and could supervise such QA models. Therefore, the key frame selector, an optional plug-and-play module for RefVOS, can be enhanced by the guidance of our \ourwork. 

Concretely, we let \ourwork annotate the Intersection over Union (IoU) scores on the training set, which functions as pseudo-labels for fine-tuning the video QA-based key frame selector to predict the IoU score on each frame. Intuitively, the key frame is simply selected as the frame with the highest score. As we noticed in Sec.~\ref{sec:ablation_self_refine_key_frame}, these pseudo-labels effectively improve the benefits of key frame selection.

\mypar{\ourwork Propagation without External VOS.} With the key frame selected as the anchor, existing methods~\cite{yan2024visa, bai2024one} utilize external VOS models to propagate the results to other frames. However, these \emph{online} modules cannot access global video contexts and comprehend the language instructions compared with our \ourwork. Therefore, we distinguish ourselves by utilizing \ourwork itself as the propagation module. Specifically, \ourwork initiates the RefVOS procedural by treating the key frame as the first query frame, then conducts RefVOS on both forward and backward directions of the video. More details are in the supplementary Sec.~\ref{sec:selector_details}. 

%% file: sec/4_exp.tex
\vspace{-2mm}
\section{Experiments}
\label{sec:exp}
\vspace{-1mm}

\subsection{Datasets and Evaluation Metrics}
\vspace{-1mm}

\mypar{MeViS.} Our study primarily focuses on the MeViS~\cite{ding2023mevis}, which presents complex video scenarios with multiple similar objects and intricate motion patterns. It is regarded as the most challenging RefVOS benchmark at present.

\mypar{Ref-Youtube-VOS.} We additionally evaluate on Ref-Youtube-VOS~\cite{seo2020urvos}, which is an earlier and widely adopted dataset. Compared with MeViS, this dataset generally exhibits simpler scenarios and language expressions focusing less on the object motions. So its challenging level is lower.

\mypar{ReVOS \& ReasonVOS.} \ourwork could also tackle various scenarios that require reasoning with the LLMs. Therefore, we also evaluate our \ourwork on Reasoning VOS~\cite{yan2024visa, zheng2024villa, bai2024one} datasets concerning reasoning abilities, \emph{i.e.}, reasoning with world knowledge.

\mypar{Evaluation Metrics.} Unless specified otherwise, the evaluation metrics used are $\mathcal{J}$ (average IoU), $\mathcal{F}$ (boundary F measure), and $\mathcal{J}\& \mathcal{F}$ (average of $\mathcal{J}$ and $\mathcal{F}$). 

\vspace{-1.0mm}
\subsection{Baselines and Implementation Details}\label{sec:implementation_details}
\vspace{-1.0mm}

\mypar{Model Architecture.} Unless specified otherwise, we adopt LISA-7B-v1~\cite{lai2024lisa} as the base MLLM to provide initial image segmentation abilities. We adopt the mask decoder from SAM-2~\cite{ravi2024sam2} as our segmentation decoder and incorporate the SAM-2 memory attention for end-to-end optimization with the memory bank. During training, only the MLLM and the SAM-2 decoder are trainable and the MLLM are fine-tuned with LoRA~\cite{hu2021lora} for efficiency. The key frame selector is fine-tuned from Chat-UniVi-7B~\cite{jin2024chat}.

\mypar{Multi-dataset Supervised Finetuning (SFT).} Existing MLLM-based RefVOS methods commonly integrate multiple datasets for training to overcome the contradiction between the scarcity of video segmentation data and the large capacities of language models. Their training sets include RefVOS, image-based questions answering and segmentation, and VOS datasets. In comparison, we initialized from LISA and conducted SFT on RefVOS. We provide two SFT options: (1) Standard-SFT (\textbf{\ourwork}$^{S}$): We utilize MeViS and Ref-Youtube-VOS for SFT, and (2) Additional-SFT (\textbf{\ourwork}$^{A}$): we further include ReVOS~\cite{yan2024visa}, LV-VIS~\cite{lvvis2023} and DAVIS-17~\cite{davis2017} to alleviate the data scarcity issue and tackle reasoning VOS. Due to limited computational resources, our Additional-SFT does not leverage object contrastive loss and keyframe selector. Notably, \ourwork utilizes fewer datasets for SFT yet achieves better performance than prior arts.

\mypar{Training Setup.} For inputs, we use a context window of $N=8$ frames, with 4 input context frames and 4 query frames. Due to our limited computation, we downsample the features of every frame by 4x, resulting in 64 visual tokens per frame. The complete training process requires $\sim$25 hours on 4 NVIDIA 40G A100 GPUs, with 3000 default optimization steps. Each step corresponds to a batch size of 2 per device and 10 gradient accumulation steps. More implementation details are in the supplementary Sec.~\ref{sec:glus_training}.

\label{sec:refvos_results}
\input{tables/main_results_v1}
\vspace{-1.5mm}
\subsection{Referring VOS Comparison}
\vspace{-1.5mm}

\mypar{MeViS and Ref-Youtube-VOS.} In Table~\ref{table:rvos_main_results}, we compare \ourwork with previous methods on two major RefVOS datasets: MeViS and Ref-Youtube-VOS. Our approach sets a new state-of-the-art on the challenging MeViS, with a substantial improvement: \ourwork handles complex video scenarios with a $\mathcal{J}\& \mathcal{F}$ boost of over 5\% compared to other MLLM-based RefVOS models. Additionally, \ourwork surpasses the previous SOTA model, DsHmp~\cite{DsHmp}, with an approximately 4\% $\mathcal{J}\& \mathcal{F}$ improvement. 

On Ref-Youtube-VOS, \ourwork outperforms most MLLM-based RefVOS models, only slightly lagging behind ViLLa~\cite{zheng2024villa} on $\mathcal{F}$ metric when using the subset of SFT datasets. Notably, previous works~\cite{yan2024visa, zheng2024villa, bai2024one}, including ViLLa, leverage more datasets spanning different tasks for SFT. Under a cleaner comparison, where our SFT similarly utilizes additional training sets (additional-SFT), our \ourwork shows better performance on Ref-Youtube-VOS.

\input{tables/reason_results}
\mypar{ReVOS \& ReasonVOS.}
We additionally evaluate our \ourwork in RefVOS scenes that require multi-modal reasoning capabilities: ReVOS and ReasonVOS. Since only our additional-SFT includes the training set of Reasoning VOS, we use GLUS$^F$ for comparison, as in Table~\ref{table:reasonvos_main_results}. \ourwork demonstrates significant improvements compared with previous MLLM VISA~\cite{yan2024visa}. In this way, our model shows consistent improvement on various reasoning tasks, \emph{e.g.}, reasoning with object motion and world knowledge, which also demonstrates the necessity of utilizing MLLMs in RefVOS.

\input{tables/ablation_full_table}

\begin{figure*}[t]
    \centering
    \includegraphics[width=0.99\textwidth]{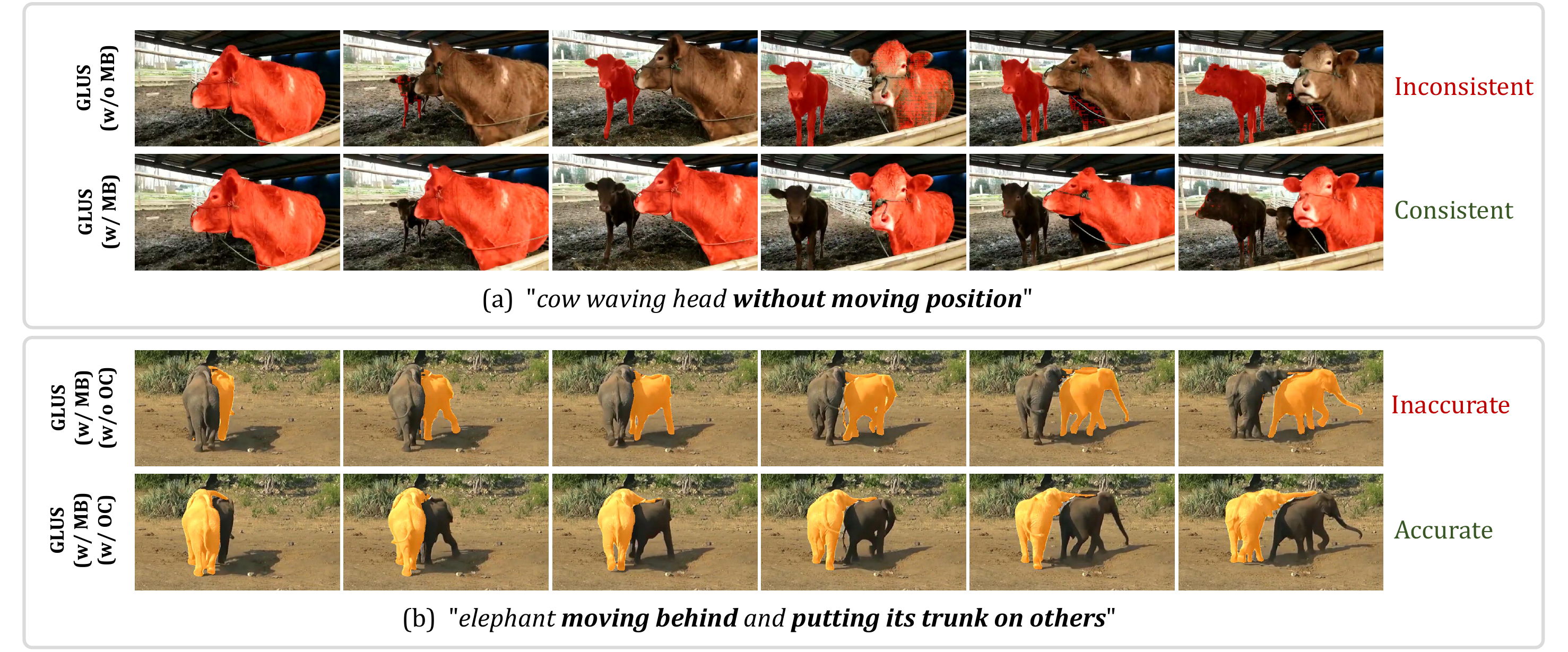}
    \vspace{-2mm}
    \caption{Qualitative comparisons on our key designs. ``MB'': End-to-end memory bank, ``OC'': Object contrastive loss. Our memory bank improves the temporal consistency of segmentation, and our object contrastive loss encourages more precise localization of objects.
    }
    \label{fig:ablation}
    \vspace{-3mm}
\end{figure*}

\vspace{-1.0mm}
\subsection{Ablation Studies}
\label{sec:ablation}
\vspace{-1mm}
\subsubsection{Global-local Unified Reasoning} We analyze the effect of our global-local unified reasoning (Sec.~\ref{sec:global-local-segmentor}) in Table~\ref{tab:ablation_full}, where either global-only or local-only reasoning performs worse than our unified strategy (``GLU''). Our qualitative results in Fig.~\ref{fig:teaser} also suggest the strength of our \ourwork.

\subsubsection{End-to-end Memory Bank} As discussed in Sec.~\ref{sec:memory_bank}, \ourwork is inherently compatible with a VOS memory bank and optimize the MLLM end-to-end with the memory bank to better comprehend historical information. As in Table~\ref{tab:ablation_full} (``GLU + MB''), \ourwork improves on both RefVOS datasets after end-to-end training and inference with SAM-2's memory attention and memory banks. Qualitative observations also suggest that our unifying MLLM with the memory bank greatly enhances the consistency of the generated masks, as in Fig.~\ref{fig:ablation}\textcolor{cvprblue}{(a)}.

\subsubsection{Object Contrastive Loss} In Sec.~\ref{sec:contrastive-loss}, we introduced an object contrastive learning pipeline to enhance the fine-grained representations of $\seg$ tokens. As in Table.~\ref{tab:ablation_full} (``GLU + MU + OC''), \ourwork enjoys improvement from the object contrastive loss, even though such contrastive learning only occurs on the MeViS data. This demonstrates the effectiveness of our design and the generalizability of learning from the challenging motion data (MeViS) for video understanding. As in Fig.~\ref{fig:ablation}\textcolor{cvprblue}{(b)}, the contrastive loss corrects the misidentified referred objects.

\subsubsection{Self-refined Key Frame Selection \& Propagation}\label{sec:ablation_self_refine_key_frame} 
\input{tables/ablation_selector}
In Sec.~\ref{sec:selector}, we propose a self-refinement framework to select key frame. To illustrate the necessity of fine-tuning selector with the pseudo-labels generated from GLUS, We compare our selector choice with other selection methods and present the results in Table~\ref{tab:ablation_selector}. As in Table~\ref{tab:ablation_selector}, our fine-tuned selector outperforms off-the-shelf methods by a significant margin. These results underscore the benefits of utilizing a grounding model (\emph{e.g.}, GLUS) to provide fine-grained supervision to video question-answering models.

Additionally, unlike previous methods, which rely on an external VOS module, we propose utilizing the internal MLLM in \ourwork for propagation. We compare propagating choices between \ourwork and two external state-of-the-art VOS models: Cutie~\cite{cheng2023putting} and SAM-2~\cite{ravi2024sam2}. As presented in Table~\ref{tab:ablation_tracker}, \ourwork outperforms both VOS methods by a significant margin, highlighting the necessity of utilizing video information and referring expressions for propagation.

To further illustrate the necessity of fine-tuning selector with the pseudo-labels generated from GLUS, We compare our selector choice with other selection methods and present the results in Table~\ref{tab:ablation_selector}. ``No selector'' refers to regular \ourwork without key frame selection, and we also adopt the off-the-shelf selectors of LLaMA-VID~\cite{li2025llama} and Chat-Univ~\cite{jin2024chat} used in VISA~\cite{yan2024visa}. As in Table~\ref{tab:ablation_selector}, our fine-tuned selector outperforms off-the-shelf methods by a significant margin. These results underscore the benefits of utilizing a grounding model (\emph{e.g.}, GLUS) to provide fine-grained supervision to video question-answering models.

\subsubsection{Propagation without External VOS}
Unlike previous methods, which rely on an external VOS module for propagation after key frame selection, we propose utilizing the internal MLLM in \ourwork for this task. We compare propagating the masks from the key frame to the whole video between \ourwork and two external state-of-the-art VOS models: Cutie~\cite{cheng2023putting} and SAM-2~\cite{ravi2024sam2}. As presented in Table~\ref{tab:ablation_tracker}, \ourwork outperforms both VOS methods by a significant margin, highlighting the necessity of utilizing global video information and referring expressions for propagation.

\input{tables/ablation_tracker}

\subsubsection{Sampling Ratio of Training Datasets} As noticed in previous works~\cite{wang2024internvideo2}, balancing the training data is critical for vision language models. We observe the same when training \ourwork with Ref-Youtube-VOS and MeViS. For this ablation, we use the \ourwork with memory bank and global-local unified reasoning enabled, and train it across different sampling ratios of the two datasets. The performance and optimization steps needed for convergence are in Table~\ref{tab:ablation_sample_rate}. For balanced performance and training efficiency, we select 1:1 as the standard sampling rate for our models.

\input{tables/ablation_sample_rate}

\subsubsection{Data Scarcity of MLLM in Video Segmentation}
\label{sec:data_scarcity}
Fine-tuning LLMs requires large amounts of data, especially for video MLLMs~\cite{wang2024internvideo2, li2024llava, wang2022internvideo}. However, video data is scarce, especially when requiring fine-grained annotations like RefVOS. With the default training steps 3000, the training of \ourwork without extended datasets averagely spans $\sim$11.6 epochs over the whole frames set, which contrasts the common 1 or 2 epochs SFT schedule for vision-language models fine-tuned with sufficient data~\cite{li2024llava, wang2024internvideo2, wang2022internvideo, jin2024chat, li2024mini, lin2024vila}.  

\begin{figure}[t]
    \centering
\includegraphics[width=0.99\linewidth]{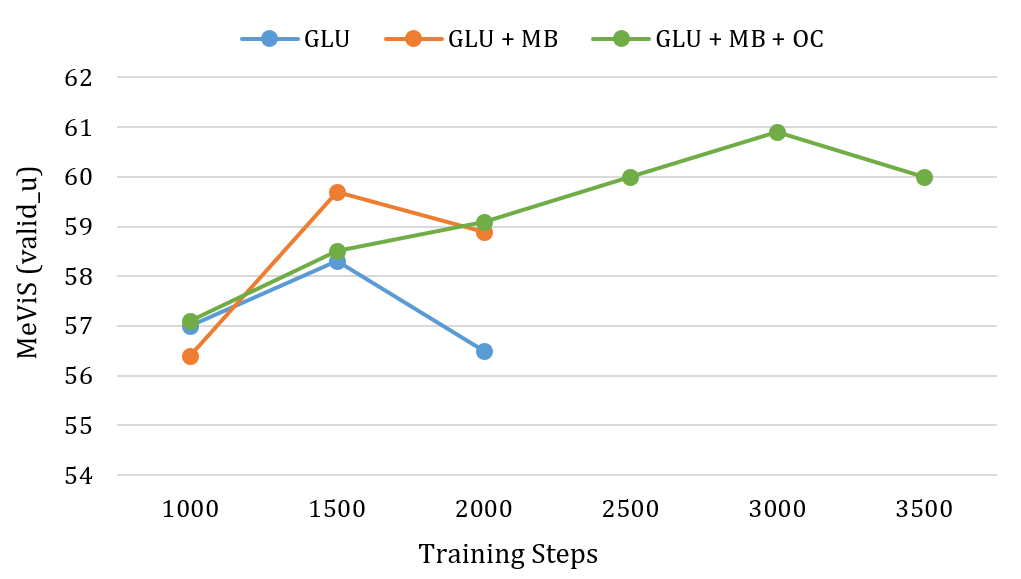}
    \vspace{-2mm}
    \caption{Curves of MeViS valid\_u performance ($\mathcal{J}$\&$\mathcal{F}$) with distinct training steps. The figure clearly demonstrates noticeable overfitting in the model. ``GLU'': Global-local unification, ``MB'': End-to-end memory bank, ``OC'': Object contrastive loss.
    }
    \label{fig:ovft}
    \vspace{-3mm}
\end{figure}

This led to noticeable overfitting with more training steps, according to the change of validation set performance (MeViS valid\_u) in Fig.~\ref{fig:ovft}. Although the object contrastive loss alleviates the overfitting issue, they all suffer from a significant drop at the final steps. We hypothesize that such a data scarcity problem constrains the performance of video MLLMs, especially when they don't have tailored designs such as hierarchical perception~\cite{DsHmp}. We hope our observation can encourage more explorations on scaling up the video segmentation data.

%% file: tables/main_results_v1.tex
\begin{table}[t]
	\centering
	\small
	\resizebox{0.49\textwidth}{!}{
		\begin{tabular}{l|c@{\hspace{2mm}}c@{\hspace{2mm}}c@{\hspace{2mm}}|c@{\hspace{2mm}}c@{\hspace{2mm}}c@{\hspace{2mm}}}
			\toprule
			\multirow{2}{*}{Method}  & \multicolumn{3}{c|}{MeViS} &\multicolumn{3}{c}{Ref-Youtube-VOS}  \\
			& $\mathcal{J}$\&$\mathcal{F}$& $\mathcal{J}$ & $\mathcal{F}$& $\mathcal{J}$\&$\mathcal{F}$ & $\mathcal{J}$ & $\mathcal{F}$ \\ 
			\midrule
            \multicolumn{7}{l}{\emph{Methods without LLMs}} \\
            \midrule
            URVOS~\cite{seo2020urvos}  &27.8 & 25.7 & 29.9 & 47.2 & 45.2 & 49.1   \\
			LBDT~\cite{Ding_2022_CVPR}  & 29.3&  27.8 & 30.8 &49.4 &48.2 &50.6 \\
            MTTR~\cite{mttr}   & 30.0 & 28.8 & 31.2 & 55.3 & 54.0 & 56.6  \\
			ReferFormer~\cite{wu2022referformer}  & 31.0 &29.8 &32.2 & 62.9 & {61.3} & 64.6 \\
            OnlineRefer~\cite{wu2023onlinerefer}  &- &- &- &63.5 &61.6 &65.5 \\
            SOC~\cite{SOC} &- &- &- & \textbf{67.3} & 65.3 & \textbf{69.3} \\
            TempCD~\cite{tang2023temporal} &- &- &- &65.8 &63.6 &68.0\\
            LoSh~\cite{yuan2024losh}&- &- &- &64.2 &62.5 &66.0 \\
            LMPM~\cite{ding2023mevis}   & 37.2 & 34.2 & 40.2 & -& -& -        \\
            DsHmp~\cite{DsHmp}  & 46.4 & 43.0 & 49.8 & 67.1 & 65.0 & 69.1                  \\
            \midrule
            \multicolumn{7}{l}{\emph{Methods with LLMs}} \\
            \midrule
			LISA-7B~\cite{lai2024lisa}  &37.2&35.1&39.4 &53.9&53.4&54.3\\
			LISA-13B~\cite{lai2024lisa}  &37.9&35.8&40.0&54.4&54.0&54.8\\
			TrackGPT-7B~\cite{zhu2023tracking}    &40.1 &37.6 &42.6 &56.4  &55.3  &57.4 \\
			TrackGPT-13B~\cite{zhu2023tracking}  &41.2  &39.2  &43.1 & 59.5 & 58.1 & 60.8\\
            VideoGLaMM~\cite{munasinghe2024videoglamm} & 45.2 & 42.1 & 48.2 & - & - & - \\
            VideoLISA-3.8B~\cite{bai2024one}   & 44.4 & 41.3 & 47.6 & 63.7 & 61.7 & 65.7 \\
            VISA-7B~\cite{yan2024visa}  & 43.5 & 40.7 & 46.3 & 61.5 & 59.8 & 63.2\\
            VISA-13B~\cite{yan2024visa}  & 44.5 & 41.8 & 47.1 & 63.0 & 61.4 & 64.7\\
            ViLLa~\cite{zheng2024villa}  & - & - & - & 66.5 & 64.6 & 68.6 \\
            \midrule
            \textbf{\ourwork}$^{S}$ (ours)  & 50.3 & 47.5 & 53.2 & 66.6 & 65.0 & 68.3 \\
            \textbf{\ourwork}$^{A}$ (ours) & \underline{\textbf{51.3}} & \underline{\textbf{48.5}} & \underline{\textbf{54.2}} & \underline{\textbf{67.3}} & \underline{\textbf{65.5}} & \underline{69.0} \\
            \bottomrule
	\end{tabular} 
        }
        \vspace{-2mm}
        \caption{The quantitative evaluation results on MeViS and Refer-Youtube-VOS. Our \ourwork performs significantly better on the most challenging MeViS benchmark, which emphasizes understanding the complex motions of objects. Meanwhile, \ourwork performs comparatively with other MLLM approaches. These support the effectiveness of our design. ``-'' means the performance not reported by a method; \textbf{bold} denotes the best scores; \underline{underline} denotes the best scores among MLLM-based methods. ``\textbf{\ourwork}$^{S}$'' and ``\textbf{\ourwork}$^{A}$'' denote the dataset options of standard-SFT and additional-SFT (Sec.~\ref{sec:implementation_details}), respectively. }
	\vspace{-3mm}
	\label{table:rvos_main_results}
\end{table}

%% file: tables/reason_results.tex
\begin{table}[t]
	\centering
	\small
	\resizebox{0.49\textwidth}{!}{
		\begin{tabular}{l|c@{\hspace{1.5mm}}c@{\hspace{1.5mm}}c@{\hspace{1.5mm}}|c@{\hspace{1.5mm}}c@{\hspace{1.5mm}}c@{\hspace{1.5mm}}|c@{\hspace{1.5mm}}|c@{\hspace{1.5mm}}c@{\hspace{1.5mm}}c@{\hspace{1.5mm}}}
			\toprule
			\multirow{3}{*}{Method}  & \multicolumn{7}{c|}{ReVOS} & \multicolumn{3}{c}{{ReasonVOS}} \\ 
            & \multicolumn{3}{c|}{Reasoning} &\multicolumn{3}{c|}{Referring} & \multirow{2}{*}{Robustness} &  \multirow{2}{*}{$\mathcal{J}$\&$\mathcal{F}$}& \multirow{2}{*}{$\mathcal{J}$} & \multirow{2}{*}{$\mathcal{F}$}\\ 
			& $\mathcal{J}$\&$\mathcal{F}$& $\mathcal{J}$ & $\mathcal{F}$& $\mathcal{J}$\&$\mathcal{F}$ & $\mathcal{J}$ & $\mathcal{F}$ & & & & \\ 
		\midrule
            VISA-7B~\cite{yan2024visa}  & 43.0 & 40.6 & 45.4 & 50.9 & 49.2 & 52.6 & 15.5 & - & - & -\\
            VISA-13B~\cite{yan2024visa}  & 44.3 & 42.0 & 46.7 & 57.4 & 55.6 & 59.1 & 14.5& - & - & - \\
            VideoLISA~\cite{bai2024one} & - & - & - & - & - & - & - & 47.5 & 45.1 & 49.9 \\
            \textbf{\ourwork}$^A$ (ours) & \textbf{51.4} & \textbf{48.8} & \textbf{53.9} & \textbf{58.3} & \textbf{56.0} & \textbf{60.7} & \textbf{17.9} & \textbf{49.9} & \textbf{47.5} & \textbf{52.4} \\
            \bottomrule
	\end{tabular} 
        }
        \vspace{-2mm}
        \caption{The quantitative evaluation results on ReVOS and ReasonVOS. Our \ourwork performs significantly better on both datasets, among both reasoning data and referring data, which demonstrates the effect of \ourwork in complex reasoning-require scenarios.}
	\vspace{-2mm}
	\label{table:reasonvos_main_results}
\end{table}

%% file: tables/ablation_full_table.tex
\begin{table}[t]
\centering
\resizebox{0.99\linewidth}{!}{
\begin{tabular}{l@{\hspace{8mm}}|c@{\hspace{4mm}}c@{\hspace{4mm}}c@{\hspace{4mm}}}
    \toprule
         \multirow{2}{*}{Method} & MeViS & MeViS & RefYTB \\
         & (valid\_u) & (valid) & (valid) \\
    \midrule
        \multicolumn{4}{l}{Baselines} \\
    \midrule
        Global Reasoning  & 55.1 & 47.2 & 62.8 \\
        Local Reasoning  & 56.9 & 46.2 & 61.6 \\
    \midrule
        \multicolumn{4}{l}{\ourwork (Ours)} \\
    \midrule
        + GLU & 58.3 & 47.4 & 63.6 \\
        + GLU + MB & 59.7 & 49.5 & 65.2 \\
        + GLU + MB + OC & 60.9 & \textbf{50.3} & 65.5\\
        + GLU + MB + OC + KFS & \textbf{61.6} & \textbf{50.3} &  \textbf{66.6} \\
    \bottomrule
\end{tabular}
}
\vspace{-2mm}
\caption{Every design from \ourwork effectively improves the performance. ``GLU'': Global-local unification (Sec.~\ref{sec:global-local-segmentor}), ``MB'': End-to-end memory bank (Sec.~\ref{sec:memory_bank}), ``OC'': Object contrastive loss (Sec.~\ref{sec:contrastive-loss}), ``KFS'': key frame selection (Sec.~\ref{sec:selector}).
}
\vspace{-3mm}
\label{tab:ablation_full}
\end{table}

%% file: tables/ablation_selector.tex
\begin{table}[t]
\centering
\resizebox{0.99\linewidth}{!}{
\begin{tabular}{l@{\hspace{4mm}}|c@{\hspace{4mm}}c@{\hspace{4mm}}c@{\hspace{4mm}}}
\toprule
 \multirow{2}{*}{Method} & MeViS & MeViS & RefYTB \\
         & (valid\_u) & (valid) & (valid) \\ 
 \midrule
GLUS No Selector &60.9 & \textbf{50.3}& 65.5\\
\midrule
LLaMA-VID 13B~\cite{li2025llama} & \textbf{61.6} &  49.8& 65.5\\
Chat-Univi 7B~\cite{jin2024chat}& 61.5  & 49.9 & 65.4 \\
Fine-tuned Chat-Univi 7B (ours)& \textbf{61.6} &  \textbf{50.3} & \textbf{66.6}\\

\bottomrule
\end{tabular}
}
\vspace{-2mm}
\caption{Ablation study on choosing key frames. We compare our selector with no selection and two baselines that utilize an off-the-shelf model for selection. We show the benefits of key frame selection and the necessity of our pseudo-label fine-tuning.
}
\vspace{-3mm}
\label{tab:ablation_selector}
\end{table}

%% file: tables/ablation_tracker.tex
\begin{table}[t]
\centering
\resizebox{0.85\linewidth}{!}{
\begin{tabular}{l@{\hspace{8mm}}|c@{\hspace{6mm}}c@{\hspace{6mm}}c@{\hspace{6mm}}}
\toprule
 \multirow{2}{*}{Method} & MeViS & MeViS & RefYTB \\
         & (valid\_u) & (valid) & (valid) \\ 
 \midrule
Cutie~\cite{cheng2023putting} &58.3 & 45.0& 64.9\\

SAM-2~\cite{ravi2024sam2} & 56.9 &  47.1& 65.5\\
\midrule
GLUS (ours)& \textbf{61.6} &  \textbf{50.3}& \textbf{66.6}\\
\bottomrule
\end{tabular}
}
\vspace{-2mm}
\caption{Ablation study on propagation modules. We compare our model with two state-of-the-art VOS methods Cutie~\cite{cheng2023putting} and SAM-2~\cite{ravi2024sam2}. The better performance of our GLUS indicates the effectiveness of our unified design in decoupling the need to use external VOS models for propagation. 
}
\vspace{-3mm}
\label{tab:ablation_tracker}
\end{table}

%% file: tables/ablation_sample_rate.tex
\begin{table}[t]
\centering
\resizebox{0.99\linewidth}{!}{
\begin{tabular}{c@{\hspace{8mm}}|c@{\hspace{6mm}}c@{\hspace{6mm}}c@{\hspace{6mm}}c@{\hspace{6mm}}}
\toprule
 \multirow{2}{*}{MeViS : RefYTB} & MeViS & MeViS & RefYTB & Best \\
 & (valid\_u) & (valid) & (valid) & Step \\
 \midrule
2:1 & \textbf{60.8} & 49.0 & 64.1 & 1500 \\
1:1 & \underline{59.7} & \underline{49.5} & 65.2 & 1500  \\
1:2 & 59.6 & 49.3 & \textbf{65.6} & 2500\\
4:15 & 59.6 & \textbf{49.9} & \underline{65.5} & 3000\\
\bottomrule
\end{tabular}
}
\vspace{-2mm}
\caption{
Ablation studies on sampling ratio of MeViS:Ref-Youtube-VOS for training. We report the performance ($\mathcal{J}$\&$\mathcal{F}$) and the training steps needed for convergence. \underline{underline} denotes the second best. We select 1:1 as the standard ratio for \ourwork to balance performance across datasets and training efficiency. (The 4:15 ratio is adopted from \cite{yan2024visa}.)
}
\vspace{-3mm}
\label{tab:ablation_sample_rate}
\end{table}

%% file: sec/5_conclusion.tex
\vspace{-2mm}
\section{Conclusion}
\label{sec:conclusion}
\vspace{-2mm}

We introduce a simple yet effective framework based on MLLMs for referring video object segmentation (RefVOS). Named ``\ourwork,'' our method establishes unified global and local reasoning in a single LLM, addressing both ``Ref'' and ``VOS'' challenges. The central design is to provide MLLM with both global (\emph{context frames}) and local (\emph{query frames}) contexts. Such unified reasoning is further enhanced with end-to-end optimized VOS memory modules to improve the consistency of \ourwork. Finally, \ourwork introduces plug-and-play \emph{object contrastive loss} and \emph{pseudo-labeling} for key frame selection, enabling the MLLM to distinguish the correct object and frame with its limited context window. Our \ourwork establishes the new state of the arts on RefVOS benchmarks. We hope our baseline can inspire more systematic studies enabling MLLMs to fine-grained video understanding.

\vspace{-3mm}
\paragraph{Acknowledgments.}
This work was supported in part by NSF Grant 2106825, NIFA Award 2020-67021-32799, Amazon, the Toyota Research Institute, the IBM-Illinois Discovery Accelerator Institute, and Snap Inc. This work used computational resources, including the NCSA Delta and DeltaAI supercomputers through allocations CIS230012 and CIS240387 from the Advanced Cyberinfrastructure Coordination Ecosystem: Services \& Support (ACCESS) program, as well as the TACC Frontera supercomputer and Amazon Web Services (AWS) through the National Artificial Intelligence Research Resource (NAIRR) Pilot.


%% file: sec/X_suppl.tex
\clearpage

\setcounter{page}{1}
\maketitlesupplementary

\renewcommand\thesection{\Alph{section}}
\renewcommand\thetable{\Alph{table}}
\renewcommand\thefigure{\Alph{figure}}
\renewcommand\theequation{\Alph{equation}}

\setcounter{section}{0}
\setcounter{table}{0}
\setcounter{figure}{0}
\setcounter{equation}{0}

\input{supp_sec/demo}
\input{supp_sec/details}

\input{supp_sec/ablation}

%% file: supp_sec/demo.tex
\section{Demo Video}

In \href{https://youtu.be/hCMah_rzVSE}{Demo}, we provide six qualitative comparisons between the previous state-of-the-art (DsHmp~\cite{DsHmp}) and our \ourwork with the videos in MeViS~\cite{ding2023mevis}. Notably, these examples illustrate three challenging aspects of RefVOS: (1) \textbf{Motion Understanding}: RefVOS models have to distinguish similar objects with their motions; (2) \textbf{Global Reasoning}: RefVOS models should be capable of realizing global reasoning to segment the objects presented only in a short video clip; (3) \textbf{Vision-Language Reasoning}: RefVOS models should perform vision-language unified reasoning in complex scenarios. The six examples demonstrate that our \ourwork effectively tackles RefVOS in challenging language-guided segmentation cases.

%% file: supp_sec/details.tex
\section{Implementation Details}

This section provides a detailed explanation of the specific model architectures and workflow of \ourwork.

\subsection{Model Architectures}

\mypar{Multimodal LLM. } The input embeddings for the MLLM are generated by processing each context and query frame individually through the vision backbone, $\mathtt{VB}$. Subsequently, a vision-to-language projection layer, $\phi_{V\to L}$, is applied to the outputs:
\begin{equation}
    \begin{split}
        F^C_t = \phi_{V\to L}(\mathtt{VB}(I^C_t)), F^Q_t = \phi_{V\to L}(\mathtt{VB}(I^Q_t)),
    \end{split}
\end{equation}
where $F_t^C$ and $F_t^Q$ are the features for the context and query frames.
Then MLLM generates the $t$-th segmentation token as:
\begin{equation}
\begin{split}
    \seg_t = \mathtt{LLM}([& R, F^C_{1:N_c}, \\
                           & F^Q_{1}, \seg_1, F^Q_{2}, \seg_2, ..., F^Q_{t}]).
\end{split}
\end{equation}

This process follows our global-local unified design, and we adopt LISA-7B-v1~\cite{lai2024lisa} for the initialization of $\mathtt{LLM}$, projector $\phi_{V\to L}$, and backbone $\mathtt{VB}$.

\mypar{Mask Decoder.} Our utilization of the mask decoder follows the style of LISA~\cite{lai2024lisa} and SAM-2~\cite{ravi2024sam2}. After obtaining $\seg_t$, \ourwork first extracts the hidden embedding $\hat{h}_t$ from the penultimate layer of the MLLM. A language-to-vision projection layer, $\phi_{L\to V}$, is then applied to $\hat{h}_t$ to generate a prompt for the mask decoder, $h_t$. Next, a vision encoder, $\mathtt{Enc}$, processes the query frames to produce encoded features. Using the prompt and the encoded features, the mask decoder, $\mathtt{Dec}$, is applied to the query image $I^Q_t$, generating its corresponding mask $M_t$:
\begin{equation}
    \begin{split}
        h_t = \phi_{L\to V}(\hat{h}_t), M_t=\mathtt{Dec}(\mathtt{Enc}(I^Q_t), h_t)
    \end{split}
\end{equation}

In our experiments, we initialize the weights of $\phi_{L\to V}$ projection layer with LISA-7B-v1 and utilize SAM-2 to initialize image encoder $\mathtt{Enc}$ and mask decoder $\mathtt{Dec}$..

\mypar{Memory Bank. } Each time a mask $M_t$ is generated, \ourwork is able to encode it using a memory encoder, $\mathtt{Enc}_M$, and stores the resulting feature $F^M_t$ in $\mathtt{MemBank}$. For memory attention, we adopt the design of SAM-2~\cite{ravi2024sam2}, selecting features from up to $m$ masks in $\mathtt{MemBank}$. Attention is then applied to these features along with the decoded image to produce the input for the mask decoder:
\begin{equation}
\begin{split}
    F^M_t &= \mathtt{Enc}_M(M_t), \enspace \mathtt{MemBank.Push}(F^M_t) \\
    \hat{F}^M_{t+1} &= \mathtt{Concat}(F^M_{i_1}, F^M_{i_2}, ..., F^M_{i_m}) \\
    \hat{F}^Q_{t+1} &= \mathtt{MemAttn}(\mathtt{Enc}(I^Q_{t+1}), \hat{F}^M_{t+1}) \\ 
    M_{t+1} &= \mathtt{Dec}(\hat{F}^Q_{t+1}, h_{t+1})
\end{split}
\end{equation}
where $\{i_p\}_{p=1}^m$ is the selected masks from memory bank following SAM-2. We adopt SAM-2's memory attention module and memory encoder in our experiments.

\subsection{GLUS Training Details} \label{sec:glus_training} This section provides detailed training configurations for \ourwork (Sec.~\textcolor{cvprblue}{4}), as summarized in Table~\ref{tab:details}. During training, only the MLLM (fine-tuned with LoRA~\cite{hu2021lora}), mask decoder, and projection layers are trainable. DeepSpeed~\cite{rasley2020deepspeed} is employed to improve training efficiency. The sampling frequency in the memory bank is set to 1 during training to maximize its utilization. The training process takes approximately 25 hours on 4 NVIDIA A100 GPUs (40 GB each), with 3000 steps, 10 gradient accumulation steps and a batch size of 2 per device.

The training objective incorporates cross entropy (CE) loss, mask loss (comprising mask DICE loss and mask BCE loss), and contrastive loss, as described in Sec.~\textcolor{cvprblue}{4.3}. The corresponding weights, $\lambda_{ce}$, $\lambda_{dice}$, $\lambda_{bce}$, and $\lambda_{ct}$, are used to compute their respective averages.

\input{supp_tables/training_details}

\subsection{GLUS Inference Details} During inference, \ourwork employs a sliding window approach with a size of 4 and a stride of 1 for the query frames. The mask of the last query frame is used as the context of the next group of query frames. The sampling frequency for the memory bank is set to sample once per 3 frames, and a maximum of 7 masks are used in mask attention. Additional ablation studies on sampling frequency are provided in Sec.~\ref{sec:ablation_sample_frequency}.

\subsection{Selector training and inference} \label{sec:selector_details}
\mypar{Data Annotation}
To generate the pseudo-labels for fine-tuning the selector model, we use \ourwork to generate the masks on the training set and compute the IoU of the masks. To mitigate the risk of overfitting, we adopt an early-stop model (trained for 500 steps) rather than the final model (trained for 3000 steps). For faster training of the selector, we label only half of the training set as the training data for selector fine-tuning.

\mypar{Implementation Details} 
We use Chat-Univi~\cite{jin2024chat} as the base Video-QA model. Similar to the design of recent grounding LLMs~\cite{lai2024lisa, yan2024visa, bai2024one, zhu2023tracking}, we introduce a special token, $\score$, into the LLM vocabulary and employ an MLP to project the corresponding embeddings. During training, we randomly sample 8 frames to represent video context and produce the score for each query frame. The hidden embedding of the score token, $\hat{h}_s$, is generated as:

\begin{equation}
\hat{h}_s = \mathtt{Selector}([P, F^C_{1:8}, F^Q, \score])
\end{equation}
where $P$ represents the language prompt. The hidden embedding of $\score$ is then projected to score $s$ through an MLP layer. The selector fine-tuning objective consists of two components: $\mathcal{L}_s$, an $L_1$ loss that supervises the frame score $s$ using the IoU pseudo-labels $y$ of the query frame, and $\mathcal{L}_{\text{txt}}$, a cross-entropy loss that supervises the text outputs of the LLM:

\begin{equation}
\begin{split}
    s &=  \phi_{\text{proj}}(\hat{h}_s), \\
    \mathcal{L}_s &=  \lvert y - s \rvert, \\
    \mathcal{L}_{\text{all}} &= \mathcal{L}_{\text{txt}} + \lambda_{s} \cdot \mathcal{L}_s
\end{split}
\end{equation}

For efficient training, the selector LLM is fine-tuned with LoRA~\cite{hu2021lora}, while the MLP layer is fully trainable. Further details on selector training are provided in Table~\ref{tab:selector_details}.
\input{supp_tables/selector_training}

\mypar{Inference and Propagation} 
The selector is trained to predict a confidence score for each frame in a test-time video, reflecting the importance of a frame with respect to the given expression. During inference, we first select the frame with the highest score as the key frame for each video-expression pair. We then use \ourwork to initiate tracking from the selected frame in both forward and backward propagation directions and iteratively generate the predictions for the entire video.

%% file: supp_tables/training_details.tex
\begin{table}[ht]
\centering
\resizebox{0.7\linewidth}{!}{
\begin{tabular}{l|c}
\toprule
Config & Value \\
\midrule
context frame num & 4 \\
question frame num & 4 \\
input resolution & 224 \\
features downsampling rate & 4 \\
optimizer & Adam \\
optimizer momentum & $\beta_1 , \beta_2 = 0.9, 0.95$ \\
optimizer weight decay & 0.0 \\
learning rate & 3e-4 \\
LoRA rank & 8 \\
$\lambda_{ce}$ & 1.0 \\
$\lambda_{dice}$ & 0.5 \\
$\lambda_{bce}$ & 2.0 \\
$\lambda_{ct}$ & 0.1 \\
batch size & 80 \\
gradient accumulation steps & 10 \\
warmup steps & 100 \\
\bottomrule
\end{tabular}
}
\vspace{-1mm}
\caption{
Implementation details of \ourwork training process.
}
\label{tab:details}
\end{table}

%% file: supp_tables/selector_training.tex
\begin{table}[ht]
\centering
\resizebox{0.7\linewidth}{!}{
\begin{tabular}{l|c}
\toprule
Config & Value \\
\midrule
context frame num & 8 \\
query frame num & 1 \\
optimizer & Adam \\
optimizer momentum & $\beta_1 , \beta_2 = 0.9, 0.95$ \\
optimizer weight decay & 0.0 \\
learning rate & 3e-4 \\
LoRA rank & 8 \\
$\lambda_{s}$ & 1.0 \\
batch size & 80 \\
gradient accumulation steps & 10 \\
MLP layer num & 3 \\
\bottomrule
\end{tabular}
}
\vspace{-1mm}
\caption{
Implementation details of selector training process.
}
\label{tab:selector_details}
\end{table}

%% file: supp_sec/ablation.tex
\section{Additional Studies}

\label{sec:ablation_sample_frequency}
\mypar{Memory Bank Sampling Frequency} The VOS memory bank is integrated into our framework and optimized end-to-end to enhance global-local reasoning capabilities in complex scenarios (Sec.~\textcolor{cvprblue}{4.2}). We evaluate the impact of memory stride in Table~\ref{tab:sample_frequency}, where a longer stride prioritizes global reasoning, while a shorter stride emphasizes local consistency. We show that GLUS performs stably with varied memory bank strides, because of its design unifying both global and local reasoning.

\input{supp_tables/sampling_frequency}

\section{Limitations and Future Works}

Our work mainly focuses on the \emph{fine-tuning} phase of a multimodal large language model for referring video object segmentation. Therefore, the visual backbone and LLM are limited in understanding the video. From this perspective, meaningful future work would start from an MLLM \emph{pre-trained} for video understanding to further enhance the motion understanding. 

In addition, our computational resources heavily constrain our \emph{context lengths} for an MLLM and limit the capability for video understanding. Concretely, we have to downsample the visual features and can only sample 4 context frames to summarize the video content, which might not cover the critical contexts if motions are happening fast. We hope combining our GLUS design with longer context windows can further unleash its potential.

Finally, we notice that the amount of data has become a bottleneck for video reasoning (Fig.~\ref{fig:ovft}). Therefore, future work can focus on improving the data scale and quality, where we hope the benefit of pseudo-labeling from GLUS can also be of use.

%% file: supp_tables/sampling_frequency.tex
\begin{table}[ht]
\centering
\resizebox{0.6\linewidth}{!}{
\begin{tabular}{l|c}
\toprule
Sampling Frequency & MeViS (valid\_u) \\
\midrule
w/o MB & 58.3 \\
\midrule
1 & 59.3 \\
3 & 59.7 \\
5 & 59.7 \\
7 & 59.7 \\
9 & 59.7 \\
\bottomrule
\end{tabular}
}
\vspace{-1mm}
\caption{
Ablation studies on the sampling frequency of memory bank. We select 3 as the default stride of the sampling frequency, following SAM2. ``MB'': Memory Bank.
}
\label{tab:sample_frequency}
\end{table}